%% file: main.tex
\documentclass[10pt,twocolumn,letterpaper]{article}
\usepackage[pagenumbers]{cvpr}

\usepackage{graphicx}
\usepackage{multicol}
\usepackage{amsmath}
\usepackage{amssymb}
\usepackage{booktabs}
\usepackage{algorithm}
\usepackage{caption}
\usepackage{subcaption}
\usepackage{algpseudocode}
\usepackage[dvipsnames]{xcolor}
\input{common_defs}

\newcommand{\FIDinf}{\operatorname{FID}_\infty}

\usepackage{pifont}%
\newcommand{\cmark}{\ding{51}}%
\newcommand{\xmark}{\ding{55}}%

\usepackage[pagebackref,breaklinks,colorlinks]{hyperref}

\usepackage[capitalize]{cleveref}
\crefname{section}{Sec.}{Secs.}
\Crefname{section}{Section}{Sections}
\Crefname{table}{Table}{Tables}
\crefname{table}{Tab.}{Tabs.}

\input{preamble}
\definecolor{cvprblue}{rgb}{0.21,0.49,0.74}

\def\confName{CVPR}
\def\confYear{2024}

\title{Rethinking FID: Towards a Better Evaluation Metric for Image Generation}

\author{
Sadeep Jayasumana \qquad Srikumar Ramalingam \qquad Andreas Veit \qquad Daniel Glasner \\
Ayan Chakrabarti \qquad Sanjiv Kumar \vspace{0.1cm} \\
Google Research, New York\\
{\tt\small \{sadeep, rsrikumar, aveit, dglasner, ayanchakrab, sanjivk\}@google.com}
}

\begin{document}
\maketitle
\input{00_abs}
\input{01_intro}
\input{02_related_works}

\input{03_body}
\input{04_exp}
\input{05_discussion}

{\small
\bibliographystyle{ieeenat_fullname}
\bibliography{refs}
}

\newpage
\input{appendix}

\end{document}

%% file: common_defs.tex
\renewcommand{\vec}[1]{\mathbf{#1}}

\usepackage{tabularx}
\usepackage{multirow}
\usepackage{amsmath}
\usepackage{amssymb}
\usepackage{graphicx}
\usepackage{xcolor}
\usepackage{colortbl}
\usepackage{stmaryrd}
\usepackage{amsthm}
\usepackage{bm}

\renewcommand{\vec}[1]{\mathbf{#1}}

\newcommand{\vx}{\vec{x}}
\newcommand{\vy}{\vec{y}}

\newcommand{\x}{\vec{x}}

\newcommand{\goodMark}{{\color{ForestGreen}\cmark}}
\newcommand{\badMark}{{\color{red}\xmark}}

%% file: 00_abs.tex
\begin{abstract}

As with many machine learning problems, the progress of image generation methods hinges on good evaluation metrics. One of the most popular is the Fr\'echet Inception Distance (FID). FID estimates the distance between a distribution of Inception-v3 features of real images, and those of images generated by the algorithm. We highlight important drawbacks of FID: Inception's poor representation of the rich and varied content generated by modern text-to-image models, incorrect normality assumptions, and poor sample complexity.
We call for a reevaluation of FID's use as the primary quality metric for generated images. We empirically demonstrate that FID contradicts human raters, it does not reflect gradual improvement of iterative text-to-image models, it does not capture distortion levels, and that it produces inconsistent results when varying the sample size.
We also propose an alternative new metric, CMMD, based on richer CLIP embeddings and the maximum mean discrepancy distance with the Gaussian RBF kernel. It is an unbiased estimator that does not make any assumptions on the probability distribution of the embeddings and is sample efficient. Through extensive experiments and analysis, we demonstrate that FID-based evaluations of text-to-image models may be unreliable, and that CMMD offers a more robust and reliable assessment of image quality. A reference implementation of CMMD is available at:
\vspace{0.1cm}
\noindent
{\small\url{https://github.com/google-research/google-research/tree/master/cmmd}.}
\end{abstract}

%% file: 01_intro.tex
\section{Introduction}

\begin{figure}[t]
\centering
\includegraphics[width=0.5\textwidth,trim={0.7cm 1cm 2cm 0cm},clip]{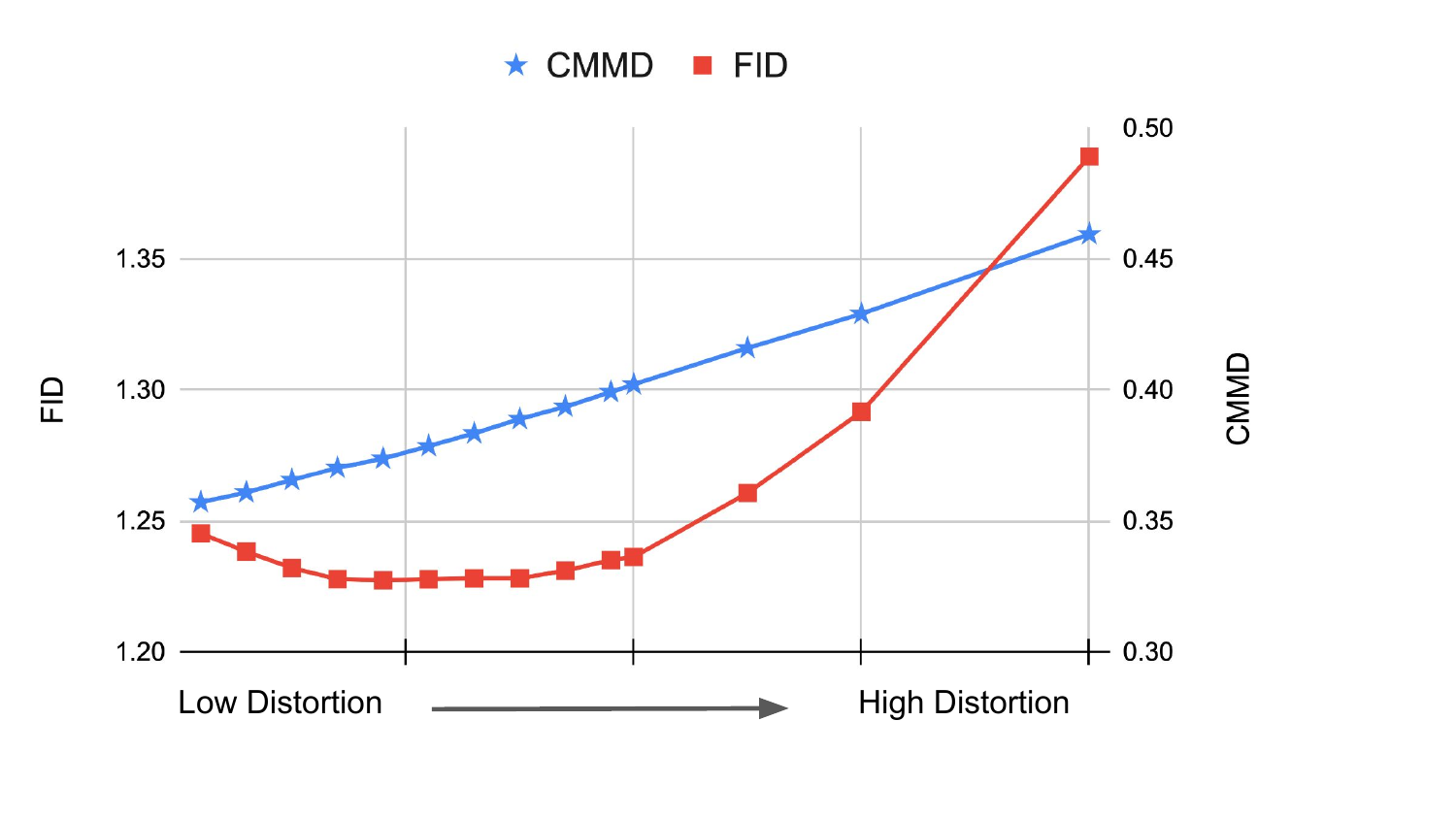}
\caption{{\it Behaviour of FID and CMMD under distortions. CMMD monotonically increases with the distortion level, correctly identifying the degradation in image quality with increasing distortions. FID is wrong. It improves (goes down) for the first few distortion levels, suggesting that quality improves when these more subtle distortions are applied. See Section~\ref{sec:vqgan_distortions} for details.}\vspace{-0.3cm}}
\label{fig:distorted_coco_images}
\end{figure}

Text-to-image models are progressing at breakneck speed. Recent models such as~\cite{saharia2022photorealistic,Rombach2022,yu2022scaling,ramesh2022hierarchical,Midjourney2022} have been incredibly successful at generating realistic images that remain faithful to text prompts. As with many problems in machine learning, a reliable evaluation metric is key to driving progress. Unfortunately, we find that the most popular metric used in the evaluation of text-to-image models, the Fr\'echet Inception Distance (FID)~\cite{heusel2018gans}, may disagree with the gold standard, human raters, in some important cases; and is thus ill-suited for this purpose. We identify some important limitations of the FID through statistical tests and empirical evaluations. To address these shortcomings, we propose an alternative metric: CMMD, which uses CLIP embeddings and Maximum Mean Discrepancy (MMD) distance. Figure~\ref{fig:distorted_coco_images} shows one of our experiments, the details of which are discussed in Section~\ref{sec:image_distortions}, in which FID does not reflect progressive distortion applied to images while CMMD correctly ranks the image sets based on the severity of the distortion. 

\begin{table*}
\begin{center}
\begin{tabular}{  l | l | l }

  &  \multicolumn{1}{c|}{Fr\'echet distance} & \multicolumn{1}{c}{MMD distance} \\
  \hline
 \multirow{4}{*}{Inception embeddings}  & \badMark~~Weak image embeddings &  \badMark~~Weak image embeddings \\
 & \badMark~~Normality assumption & \goodMark~~Distribution-free \\
 & \badMark~~Sample inefficient & \goodMark~~Sample efficient \\
 & \badMark~~Biased estimator & \goodMark~~Unbiased estimator \\
 \hline
  \multirow{4}{*}{CLIP embeddings}  & \goodMark~~Rich image embeddings &  \goodMark~~Rich image embeddings \\
 & \badMark~~Normality assumption & \goodMark~~Distribution-free \\
 & \badMark~~Sample inefficient & \goodMark~~Sample efficient \\
 & \badMark~~Biased estimator & \goodMark~~Unbiased estimator \vspace{-0.4cm}
\end{tabular}
\end{center}
\caption{\it Comparison of options for comparing two image distributions. FID, the current de facto standard for text-to-image evaluation is in the upper-left corner. The proposed metric, CMMD, is in the lower-right corner and has many desirable properties over FID.}
\label{table:metric_options}
\end{table*}

Evaluating image generation models is a uniquely challenging task. Unlike traditional vision tasks such as classification or detection, we need to evaluate multiple dimensions of performance including quality, aesthetics and faithfulness to the text prompt. Moreover, these are hard-to-quantify concepts which depend on human perception. As a result, human evaluation remains the gold standard for text-to-image research.
Since human evaluation is an expensive solution that does not scale well, researchers often rely on automated evaluation. Specifically, recent works have used FID and CLIP distance to measure image quality and faithfulness to the text prompts, respectively.

In this work, we call for a reevaluation of this approach, in particular, the use of FID as a measure of image quality. 
We highlight drawbacks of FID, such as incorrectly modeling Inception embeddings of image sets as coming from a multivariate normal distribution and its inconsistent results when varying the sample size (also noted in~\cite{Chong2019}). We empirically show that, FID can contradict human raters, does not reflect gradual improvement of iterative text-to-image models and does not capture complex image distortions.

Our proposed metric uses CLIP embeddings and the MMD distance. Unlike Inception embeddings, which were trained on about 1 million ImageNet images, restricted to $1000$ classes~\cite{SzegedyVISW15}, CLIP is trained on 400 million images with corresponding text descriptions~\cite{radford2021clip}, making it a much more suitable option for the rich and diverse content generated by modern image generation models and the intricate text prompts given to modern text-to-image models.

MMD, is a distance between probability distributions that offers some notable advantages over the Fr\'echet distance. When used with an appropriate kernel, MMD is a metric that does not make any assumptions about the distributions, unlike the Fr\'echet distance which assumes multivariate normal distributions. As shown in~\cite{Chong2019}, FID is a biased estimator, where the bias depends on the model being evaluated. MMD, on the other hand, is an unbiased estimator, and as we empirically demonstrate it does not exhibit a strong dependency on sample size like the Fr\'echet distance. Finally, it admits a simple parallel implementation. The ability to estimate from a smaller sample size and the fast computation make MMD fast and useful for practical applications. Different options for comparing two image distributions are compared in Table~\ref{table:metric_options}. The existing FID metric is in the upper-left corner and has many unfavorable properties. Our proposed metric, CMMD, is in the lower-right corner and avoids the drawbacks of FID.

\noindent
We summarize our contributions below:

\begin{itemize}

\item We call for a reevaluation of FID as the evaluation metric for modern image generation and text-to-image models. We show that it does not agree with human raters in some important cases, that it does not reflect gradual improvement of iterative text-to-image models and that it does not capture obvious image distortions.

\item We identify and analyze some shortcomings of the Fr\'echet distance and of Inception features, in the context of evaluation of image generation models.

\item We propose CMMD, a distance that uses CLIP features with the MMD distance as a more reliable and robust alternative, and show that it alleviates some of FIDs major shortcomings.

\end{itemize}

%% file: 02_related_works.tex
\section{Related Works}

\begin{table*}
\centering
\begin{tabular}{lccccccc}
   \toprule
    &
    \includegraphics[width=0.11\textwidth]{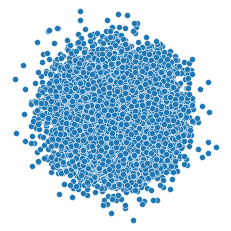} & \includegraphics[width=0.11\textwidth]{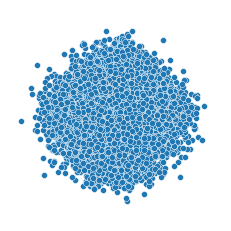} & \includegraphics[width=0.11\textwidth]{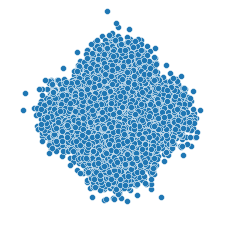} & \includegraphics[width=0.11\textwidth]{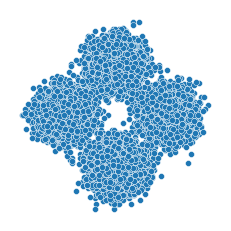} &
    \includegraphics[width=0.11\textwidth]{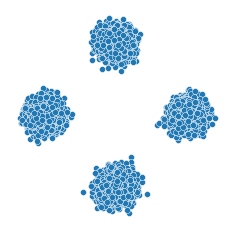} &
    \includegraphics[width=0.11\textwidth]{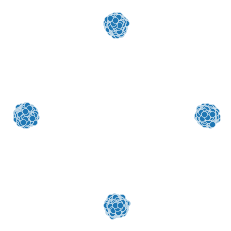} &
    \includegraphics[width=0.11\textwidth]{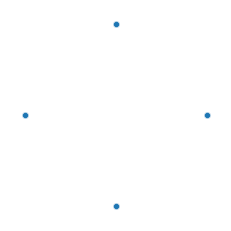} \\
    \midrule
    FD & 0.0 & 0.0 & 0.0 & 0.0 & 0.0 & 0.0 & 0.0 \\
    $\text{FD}\infty$ & 0.0 & 0.0 & 0.0 & 0.0 & 0.0 & 0.0 & 0.0 \\
    MMD & 0.0 & 0.5875 & 5.794 & 17.21 & 78.88 & 202.8 & 244.9 \\
    \bottomrule
\end{tabular}
\caption{\it Behavior of estimated Fr\'echet distances and MMD when normality assumption is violated. Going from left to right, the probability distribution changes more and more from the leftmost distribution. However, the Fr\'echet distances to the leftmost distribution calculated with normality assumption remains misleadingly zero. MMD, on the other hand, is able to correctly capture the progressive departure.
}
\label{tbl:mog_results}
\end{table*}

Generated image quality has been assessed using a variety of metrics including  log-likelihood~\cite{goodfellow2014generative}, Inception Score (IS)~\cite{SalimansGZCRC16,barratt2018note}, Kernel Inception Distance (KID)~\cite{bibkowski2021demystifying,Xu2018}, F\'rechet Inception Distance (FID) ~\cite{heusel2018gans}, perceptual path length~\cite{Karras2018}, Gaussian Parzen window~\cite{goodfellow2014generative}, and HYPE~\cite{Zhou2019}. 

IS is calculated using the Inception-v3 model~\cite{SzegedyVISW15}, which has been trained on ImageNet, to measure the diversity and quality of generated images by leveraging the 1000 class probabilities of the generated images. While IS does not require the original real images, KID and FID are computed by determining the distance between the distributions of real and generated images. KID utilizes the squared MMD distance with the rational
quadratic kernel. FID employs the squared Fr\'echet distance between two probability distributions, which is also equal to the Wasserstein-2 distance, with the assumption that both distributions are multivariate normal. Both FID and KID suffer from the limitations of the underlying Inception embeddings: they have been trained on only 1 million images, limited to 1000 classes. Intuitively, we expect this could limit their ability to represent the rich and complex image content seen in modern generated images.

Previous work has pointed to the unreliability of evaluation metrics in image generation~\cite{Chong2019, parmar2021cleanfid}. Chong et al.~\cite{Chong2019} show that FID is a biased estimator and that the bias depends on the model being evaluated. They propose an extrapolation approach to compute a bias-free estimator: $\FIDinf$. Parmar et al.~\cite{parmar2021cleanfid} show that low-level image processing operations such as compression and resizing can lead to significant variations in FID, and advocate the use of anti-aliased resizing operations. In this work, we show that FID's issues extend well beyond what is discussed in those prior works and that $\FIDinf$ and/or anti-aliased resizing do not solve those issues.

%% file: 03_body.tex
\section{Limitations of FID}

In this section we highlight some key limitations of FID. We start with a background discussion of the metric in order to better understand its limitations. Fr\'echet Inception Distance (FID) is used to measure the discrepancy between two image sets: $\mathcal{I}$ and $\mathcal{I'}$. Usually one set of images are real (for example, from the COCO dataset) and the other set is generated using the image generation model to be evaluated. To calculate FID, Inception-v3\footnote{Throughout the paper we use the terms Inception and Inception-v3 interchangeably.} embeddings~\cite{SzegedyVISW15} are first extracted for both image sets using the Inception-v3 model trained on the ImageNet classification task. The FID between $\mathcal{I}$ and $\mathcal{I'}$ is then defined as the Fr\'echet distance between these two sets of Inception embeddings.

\subsection{The Fr\'echet Distance}

For any two probability distributions $P$ and $Q$ over $\mathbb{R}^d$ having finite first and second moments, the Fr\'echet distance is defined by~\cite{Frechet1957, Dowson1982}:

\begin{equation}
    \operatorname{dist}^2_F (P, Q) := \inf_{\gamma \in \Gamma(P, Q)} \mathbb{E}_{(\vx, \vy) \sim \gamma} \|\vx - \vy\|^2,
\end{equation}
where $\Gamma(P, Q)$ is the set of all couplings of P and Q. This is also equivalent to the Wasserstein-2 distance on $\mathbb{R}^d$. In general, obtaining a closed-form solution for the  Fr\'echet distance is difficult. However, the authors of~\cite{Dowson1982} showed that a closed-form solution exists for multivariate normal distributions in the form:
\begin{equation}
    \operatorname{dist}_F^2 (P, Q) = \|\boldsymbol\mu_P - \boldsymbol\mu_Q\|^2_2 + \operatorname{Tr}(\boldsymbol\Sigma_P + \boldsymbol\Sigma_Q - 2(\boldsymbol\Sigma_P\boldsymbol\Sigma_Q)^{\frac{1}{2}}),
    \label{eqn:fd_for_gaussians}
\end{equation}
where $\boldsymbol\mu_P, \boldsymbol\mu_Q$ are the means and $\boldsymbol\Sigma_P, \boldsymbol\Sigma_Q$ are the covariances of the two multivariate normal distributions $P$ and $Q$. Note that this simplified formula is strictly valid only when both $P$ and $Q$ are multivariate normal distributions~\cite{Dowson1982}.

For FID, we need to estimate the Fr\'echet distance between two distributions of Inception embeddings, using two corresponding samples. This is challenging due to the high dimensionality of inception embeddings, $d = 2048$. Assuming that the Inception embeddings are drawn from a normal distribution simplifies the problem, allowing us to use Eq.~\eqref{eqn:fd_for_gaussians} with $\boldsymbol\mu_P, \boldsymbol\mu_Q$ and $\boldsymbol\Sigma_P, \boldsymbol\Sigma_Q$ estimated from the two samples $\mathcal{I}$ and $\mathcal{I'}$. There are two kinds of error in this procedure:

\begin{enumerate}
\item As we show in Section~\ref{sec:normality_tests}, Inception embeddings for typical image sets are far from being normally distributed. The implications of this inaccurate assumption when calculating the Fr\'echet distance are discussed in Section~\ref{sec:mog_toy_example}.

\item Estimating $(2048 \times 2048)$-dimensional covariance matrices from a small sample can lead to large errors, as discussed in Section~\ref{sec:sample_efficiency}.
\end{enumerate}

\subsection{Implications of Wrong Normality Assumptions}
\label{sec:mog_toy_example}
When calculating the Fr\'echet distance between two distributions, making an incorrect normality assumption can lead to disastrous results. We illustrate this using a 2D isotropic Gaussian distribution at the origin as the reference distribution and by measuring the distance between that and a series of mixture-of-Gaussian distributions generated as described below. The results are summarized in Table~\ref{tbl:mog_results}.

To generate the series of second distributions, we start with a mixture of four Gaussians, each having the same mean and covariance as the reference Gaussian. Since this mixture has the same distribution as the reference distribution, we expect any reasonable distance to measure zero distance between this and the reference distribution (first column of Table~\ref{tbl:mog_results}). We then let the second distribution's four components get further and further away from each other while keeping the overall mean and the covariance fixed (first row of Table~\ref{tbl:mog_results}). When this happens the second distribution obviously gets further and further away from the reference distribution. However, the Fr\'echet distance calculated with the normality assumption (note that this is \emph{not} the true Fr\'echet distance, which cannot be easily calculated) remains misleadingly zero. This happens because the second distribution is normal only at the start, therefore the normality assumption is reasonable only for the first column of the table. Since the second distribution is not normal after that, the Fr\'echet distance calculated with normality assumption gives completely incorrect results. Note that, as shown in the third row of Table~\ref{tbl:mog_results}, $\FIDinf$, the unbiased version of FID proposed in~\cite{Chong2019}, also suffers from this shortcoming, since it also relies on the normality assumption. In contrast, the MMD distance described in Section~\ref{sec:mmd} (bottom row of Table~\ref{tbl:mog_results}) is able to capture the progressive departure of the second distribution from the reference distribution. More details of the experiment setup are in Appendix \ref{appendix:mog_details}.

\subsection{Incorrectness of the Normality Assumption}
\label{sec:normality_tests}
When estimating the Fr\'echet distance, it is assumed that the Inception embeddings for each image set (real and generated), come from a multivariate normal distribution. In this section, we show that this assumption is wrong. As discussed in Section~\ref{sec:mog_toy_example}, making a wrong normality assumption about the underlying distribution can lead to completely wrong results.

It should not be surprising that Inception embeddings for a typical image set do not have a multivariate normal distribution with a single mode. Inception embeddings are activations extracted from the penultimate layer of the Inception-v3 network. During training, these activations are classified into one of 1000 classes using a \emph{linear} classifier (the last fully-connected layer of the Inception-v3 network). Therefore, since the Inception-v3 network obtains good classification results on the ImageNet classification task, one would expect Inception embeddings to have at least $1,000$ clusters or modes. If this is the case, they cannot be normally distributed. 

\begin{figure}[t]
\centering
\includegraphics[width=0.5\textwidth]{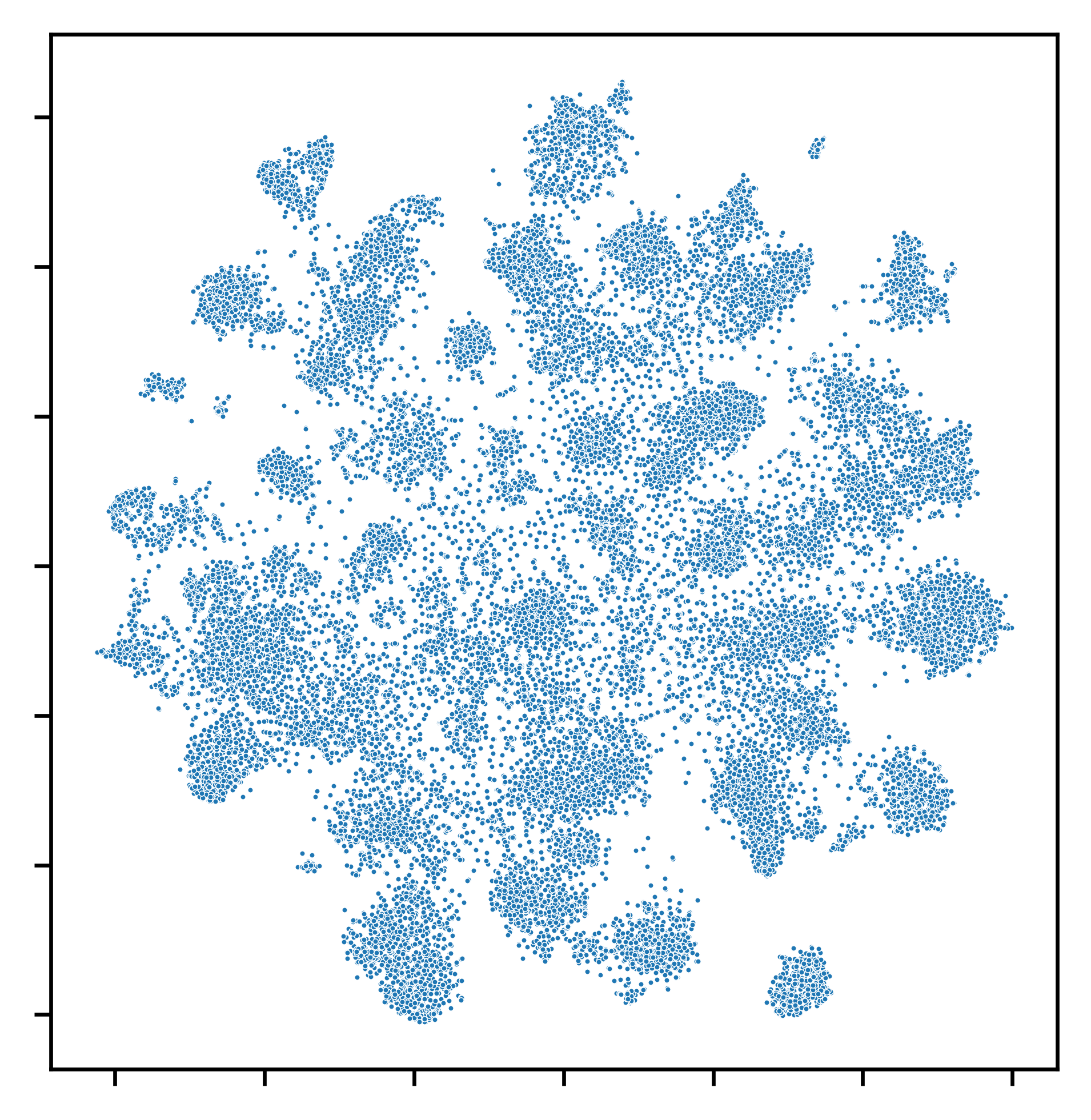}
\caption{{\it t-SNE visualization of Inception embeddings of the COCO 30K dataset. Note that even in the reduced-dimensional 2D representation, it is easy to identify that embeddings have multiple modes and do not follow a multivariate normal distribution.}\vspace{-0.3cm}}
\label{fig:inception_tsne}
\end{figure}

Figure~\ref{fig:inception_tsne} shows a 2-dimensional t-SNE~\cite{tsne} visualization of Inception embeddings of the COCO 30K dataset, commonly used as the reference (real) image set in text-to-image FID benchmarks. It is clear that the low dimensional visualization has multiple modes, and therefore, it is also clear that the original, 2048-dimensional distribution is not close to a multivariate normal distribution. 

Finally, we applied three different widely-accepted statistical tests: Mardia's skewness test, Mardia's kurtosis test, and Henze-Zirkler test to test normality of Inception embeddings of the COCO 30K dataset. All of them \emph{strongly} refute the hypothesis that Inception embeddings come from a multivariate normal distribution, with $p$-values of virtually zero (indicating an overwhelming confidence in rejecting the null hypothesis of  normality). The details of these tests can be found in Appendix~\ref{appendix:normality_tests}. 

To be clear, we do not expect CLIP embeddings to be normally distributed either. It is FID's application of Fr\'echet distance with its normality assumption to non-normal Inception features, that we object to. In fact, CLIP embeddings of COCO 30K also fail the normality tests with virtually zero $p$-values, indicating that it is not reasonable to assume normality on CLIP embeddings either.

\section{The CMMD Metric}
\label{sec:mmd}
In this section, we propose a new metric to evaluate image generation models, using CLIP embeddings and the Maximum Mean Discrepancy (MMD) distance, with a Gaussian RBF kernel. The CMMD (stands for CLIP-MMD) metric is the squared MMD distance between CLIP embeddings of the reference (real) image set and the generated image set.

CLIP embeddings~\cite{radford2021clip} have changed the way we think about image and text representations by learning them in a joint space. CLIP trains an image encoder and a text encoder jointly using 400 million image-text pairs containing complex scenes. In contrast, Inception-v3 is trained on ImageNet, which has on the order of 1 million images which are limited to 1000-classes and only one prominent object per image. As a result, CLIP embeddings are better suited for representing the diverse and complex content we see in images generated by modern image generation algorithms and the virtually infinite variety of prompts given to text-to-image models.

To compute the distance between two distributions we use the MMD distance~\cite{Gretton_NIPS2006, Gretton_2012}. MMD was originally developed as a part of a two-sample statistical test to determine whether two samples come from the same distribution. The MMD statistic calculated in this test can also be used to measure the discrepancy between two distributions. For two probability distributions $P$ and $Q$ over $\mathbb{R}^d$, the MMD distance with respect to a positive definite kernel $k$ is defined by:
\begin{align}
    \operatorname{dist}^2_{\operatorname{MMD}}(P, Q) := \mathbb{E}_{\vx, \vx'}[k(\vx, \vx')] &+ \mathbb{E}_{\vy, \vy'}[k(\vy, \vy')] \nonumber  \\
    &- 2 \mathbb{E}_{\vx, \vy}[k(\vx, \vy)],
\end{align}
where $\vx$ and $\vx'$ are independently distributed by $P$ and $\vy$ and $\vy'$ are independently distributed by $Q$. It is known that the MMD is a metric for characteristic kernels $k$~\cite{NIPS2008_Fukumizu, Gretton_2012}.

 Given two sets of vectors , $X = \{\vx_1, \vx_2, \dots, \vx_m\}$ and $ Y = \{\vy_1, \vy_2, \dots, \vy_n\}$, sampled from $P$ and $Q$, respectively, an unbiased estimator for $d^2_{\operatorname{MMD}}(P, Q)$ is given by,
\begin{align}
    \hat{\operatorname{dist}}_{\operatorname{MMD}}^2(X, Y) =& \frac{1}{m(m-1)}\sum_{i=1}^m\sum_{j \ne i}^m k(\vx_i,\vx_j) \nonumber \\
    &+ \frac{1}{n(n-1)}\sum_{i=1}^n\sum_{j \ne i}^n k(\vy_i,\vy_j) \nonumber \\
    &- \frac{2}{mn}\sum_{i=1}^m\sum_{j=1}^n k(\vx_i, \vy_j).
    \label{eqn:mmd_from_a_sample}
\end{align}

Some advantages of MMD over the Fr\'echet distance are:

\begin{enumerate}
    \item MMD metric, when used with a characteristic kernel~\cite{NIPS2008_Fukumizu}, is \emph{distribution-free}. That is, it does not make any assumptions about the distributions $P$ and $Q$. In contrast, the Fr\'echet distance in Eq.~\eqref{eqn:fd_for_gaussians} assumes normality and is liable to give erroneous results when this assumption is violated.
    
    \item As shown in \cite{Chong2019}, the FID estimated from a finite sample has a bias that depends on the model being evaluated, to the extent that the sample size can lead to different rankings of the models being evaluated. Removing this bias requires a computationally expensive procedure involving computation of multiple FID estimates~\cite{Chong2019}. In contrast, the MMD estimator in Eq.~\eqref{eqn:mmd_from_a_sample}, is \emph{unbiased}.
    
    \item When working with high-dimensional vectors such as image embeddings, MMD is \emph{sample efficient}. Fr\'echet distance, on the other hand, requires a large sample to reliably estimate the $d \times d$ covariance matrix. This will be further elaborated on in Section~\ref{sec:sample_efficiency}.

\end{enumerate}
As the kernel in the MMD calculation, we use the Gaussian RBF kernel $k(\vx, \vy) = \exp(-\|\vx - \vy\|^2/2\sigma^2)$, which is a characteristic kernel,  with the bandwidth parameter set to $\sigma=10$. Empirically, we observed that the bandwidth parameter does not significantly affect the overall trends of the metric. However, we propose to keep it fixed at $10$ to obtain consistent values for the metric. Since the MMD metric with the Gaussian kernel is bounded above at $2$ (when the two distributions are maximally different), it gives small values for general distributions. We therefore scale up the value in Eq.~\eqref{eqn:mmd_from_a_sample} by $1000$ to obtain more readable values. For the CLIP embedding model, we use the publicly-available ViT-L/14@336px model, which is the largest and the best performing CLIP model~\cite{radford2021clip}. Also note that we have $m = n$ in Eq.~\eqref{eqn:mmd_from_a_sample} for text-to-image evaluation since we evaluate generated images against real images sharing the same captions/prompts. Our code for computing CMMD is publicly available.

%% file: 04_exp.tex
\section{Human Evaluation}

\begin{figure*}[t]
    \centering
    \begin{subfigure}[t]{0.24\textwidth}
        \centering
        \includegraphics[width=\textwidth]{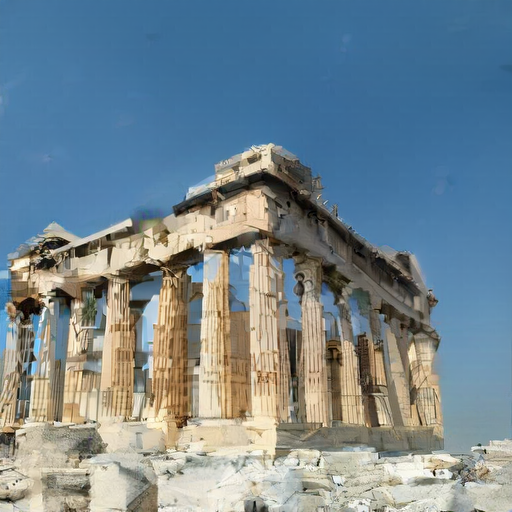}
        \caption{Step 1}
    \end{subfigure}
    \begin{subfigure}[t]{0.24\textwidth}
        \centering
        \includegraphics[width=\textwidth]{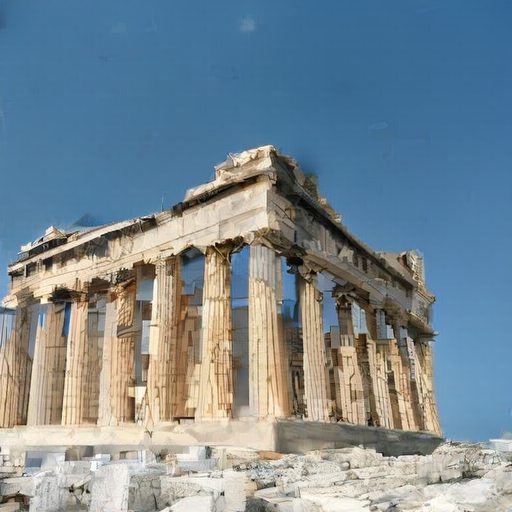}
        \caption{Step 3}
    \end{subfigure}
    \begin{subfigure}[t]{0.24\textwidth}
        \centering
        \includegraphics[width=\textwidth]{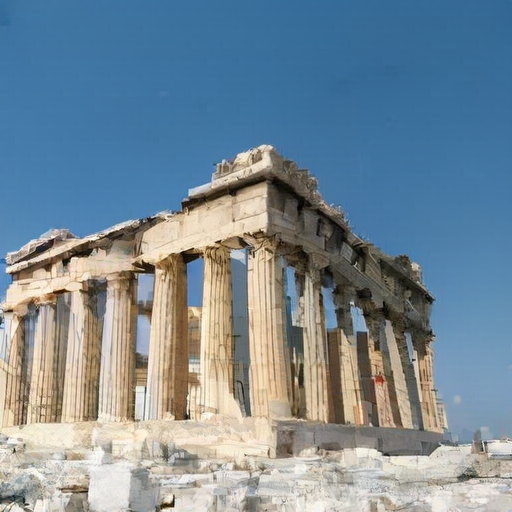}
        \caption{Step 6}
    \end{subfigure}
    \begin{subfigure}[t]{0.24\textwidth}
        \centering
        \includegraphics[width=\textwidth]{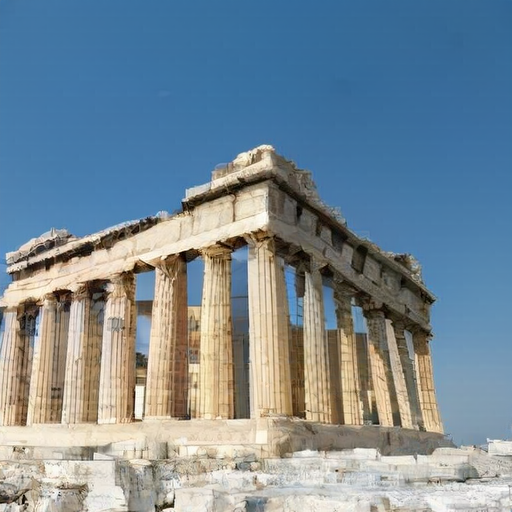}
        \caption{Step 8}
    \end{subfigure}
    \caption{{\it The quality of the generated image monotonically improves as we progress through Muse's refinement iterations. CMMD correctly identifies the improvements. FID, however, incorrectly indicates a quality degradation (see Figure~\ref{fig:muse_iterations}). Prompt: ``The Parthenon".}}
    \label{fig:muse_steps_sample}
\end{figure*}

\begin{table}[b]
    \centering
    \begin{tabular}{l r r}
    \toprule
       \multicolumn{1}{c}{Model} & \multicolumn{1}{c}{Model-A} & \multicolumn{1}{c}{Model-B} \\
        \midrule
        FID  & 21.40 & 18.42   \\
        $\FIDinf$ & 20.16 & 17.19 \\
        CMMD & 0.721 & 0.951  \\
        Human rater preference & 92.5\%   &  6.9\% \\
        \bottomrule
    \end{tabular}
    \vspace{5pt}
    \caption{\it Human evaluation of different models. FID contradicts human evaluation while CMMD agrees.}
    \label{tab:human_rating}
\end{table}

\begin{figure}[t!]
\centering
\includegraphics[width=0.48\textwidth,trim={0.7cm 0.6cm 0cm 0cm},clip]{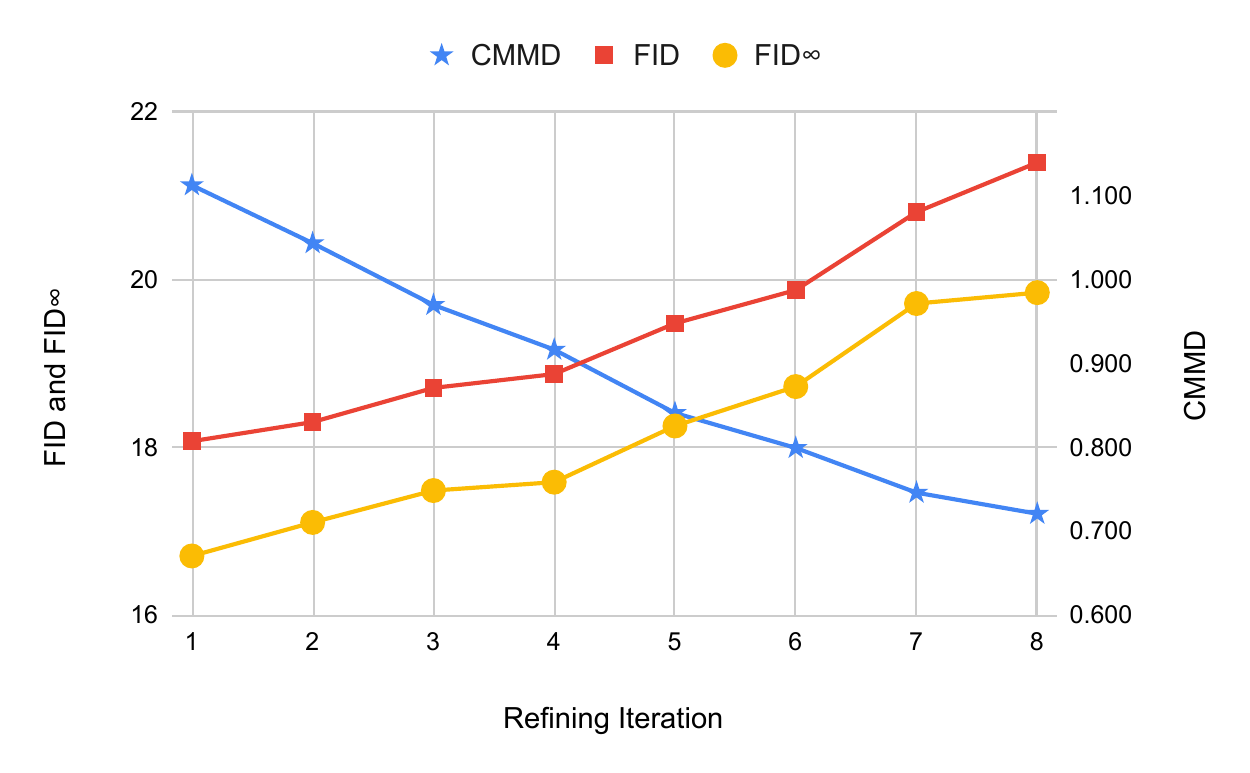}
\caption{{\it  Behavior of FID and CMMD for Muse steps. CMMD monotonically goes down, correctly identifying the iterative improvements made to the images (see Figure~\ref{fig:muse_steps_sample}). FID is completely wrong suggesting degradation in image quality as iterations progress. $\text{FID}_\infty$ has the same behavior as FID.}}
\label{fig:muse_iterations}
\end{figure}

We now present a human evaluation to show that FID does not agree with human perception of image quality. To this end, we picked two models, Model-A: the full Muse model as described in~\cite{chang2023Muse} with 24 base-model iterations and 8 super-resolution model iterations. Model-B: an early-stopped Muse model with only 20 base-model iterations and 3 super-resolution model iterations. This was done intentionally to reduce the quality of produced images.  We use a Muse model trained on the WebLI dataset~\cite{Pali2022}, generously made available to us by the Muse authors. The choice of  early-stopping iterations is arbitrary: as shown in Figure~\ref{fig:muse_iterations}, FID is consistently better (lower) for all early-stopped models when compared with the full model (Model-A).

We performed a side-by-side evaluation where human raters were presented with two images, one generated from Model-A and the other generated from Model-B. We used the same random seeds to ensure that image content and degree of alignment to the prompt are the same. This allowed the raters to focus on image quality. The raters were asked to evaluate which image looked better. Raters had the option of choosing either image or that they are indifferent. All image pairs were rated by 3 independent raters, hired through a high-quality crowd computing platform. The raters were not privy to the details of the image sets and rated images purely based on the visual quality. The authors and the raters were anonymous to each other.

We used all PartiPrompts~\cite{yu2022scaling},
which is a collection of 1633 prompts designed for text-to-image model evaluation. These prompts cover a wide range of categories (abstract, vehicles, illustrations, art, world knowledge, animals, outdoor scenes, etc.) and challenge levels (basic, complex  fine-grained detail, imagination, etc.). Evaluation results are summarized in Table~\ref{tab:human_rating}. For each comparison, we consider a model as the winner if 2 or more raters have preferred the image produced by that model. If there is no consensus among the raters or if the majority of the raters selected are indifferent, no model wins. We observed that Model-A was preferred in 92.5\% of the comparisons, while Model-B was preferred only 6.9\% of the time. The raters were indifferent 0.6\% of the time. It is therefore clear that human raters overwhelmingly prefer Model-A to Model-B. However, COCO 30K FID and its unbiased variant $\FIDinf$, unfortunately say otherwise. On the other hand, the proposed CMMD metric correctly aligns with the human preference.

\vspace{-0.1cm}
\section{Performance Comparison}\label{sec:experiments}

\begin{figure*}[h]
\centering
\includegraphics[width=0.9\textwidth,trim={0cm 7cm 7.5cm 0},clip]{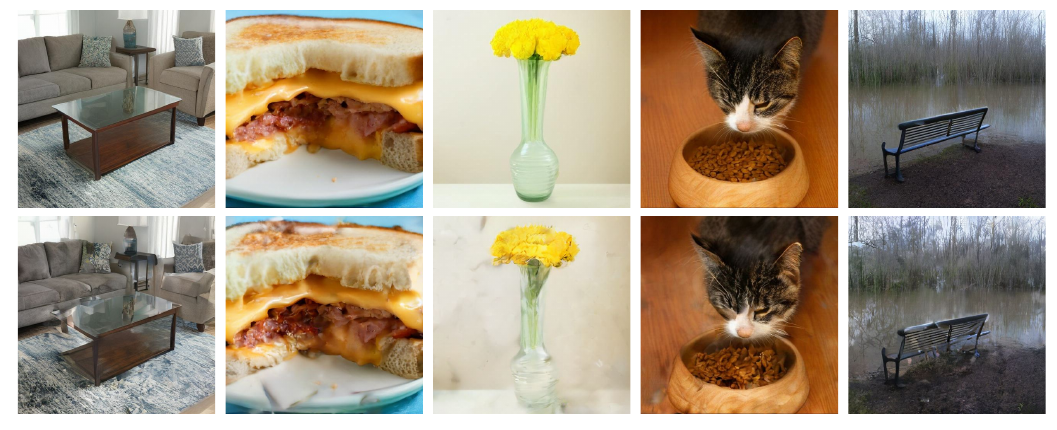}
\caption{{\it Behavior of FID and CMMD under distortions. Images in the first row (FID: 21.40, CMMD: 0.721) are undistorted. Images in the second (FID: 18.02, CMMD: 1.190) are distorted by randomly replacing each VQGAN token with probability $p=0.2$. The image quality clearly degrades as a result of the distortion, but FID suggests otherwise, while CMMD correctly identifies the degradation.} }
\label{fig:image_vis_muse_distortions}
\end{figure*}

We now compare FID with the proposed CMMD metric under various settings to point out the limitations of FID while highlighting the benefits of CMMD. In all our experiments, we use the COCO 30K dataset~\cite{lin2014microsoft} as the reference (real) image dataset. Zero-shot evaluation on this dataset is currently the de facto evaluation standard for text-to-image generation models~\cite{Rombach2022, saharia2022photorealistic, chang2023Muse}. Throughout our experiments, where applicable, we use high-quality bicubic resizing with anti-aliasing as suggested in~\cite{parmar2021cleanfid}. This prevents any adverse effects of improperly-implemented low level image processing operations on FID as those reported in~\cite{parmar2021cleanfid}.

 For Stable Diffusion~\cite{Rombach2022}, we use the publicly available Stable Diffusion 1.4 model. We evaluate all models without any additional bells and whistles such as CLIP sorting.

\subsection{Progressive Image Generation Models}
Most modern text-to-image generation models are iterative. For example, diffusion models~\cite{saharia2022photorealistic,Rombach2022} require multiple denoising steps to generate the final image, the Parti model~\cite{yu2022scaling} auto-regressively generates image tokens one at a time. While the Muse model~\cite{chang2023Muse} generates multiple tokens at a time, it still requires iterative sampling steps to generate the final image, as shown in Figure~\ref{fig:muse_steps_sample}. Gradually improving the quality of the generated images in each step, these methods go from poor quality images or pure noise images to unprecedented photo-realism. This progression in quality is obvious to a human observer and we would expect any reasonable metric to monotonically improve as we progress through iterations of image generation.

Figure~\ref{fig:muse_iterations} shows FID, $\FIDinf$, and CMMD values for progressive Muse iterations. 
FID and $\FIDinf$ incorrectly suggest that the image quality degrades, when the quality improvements are obvious as illustrated in Figure~\ref{fig:muse_steps_sample}. In contrast, CMMD correctly identifies the quality improvements made during Muse's iterative refinements. As seen in Figure~\ref{fig:muse_iterations}, we consistently observe in our experiments that FID and $\FIDinf$ have the same behavior although absolute values are different. This is not surprising since $\FIDinf$ is derived from FID and inherits many of its shortcomings. 

\begin{figure}[t!]
\centering
\includegraphics[width=0.48\textwidth,trim={0.75cm 0.6cm 0cm 0cm},clip]{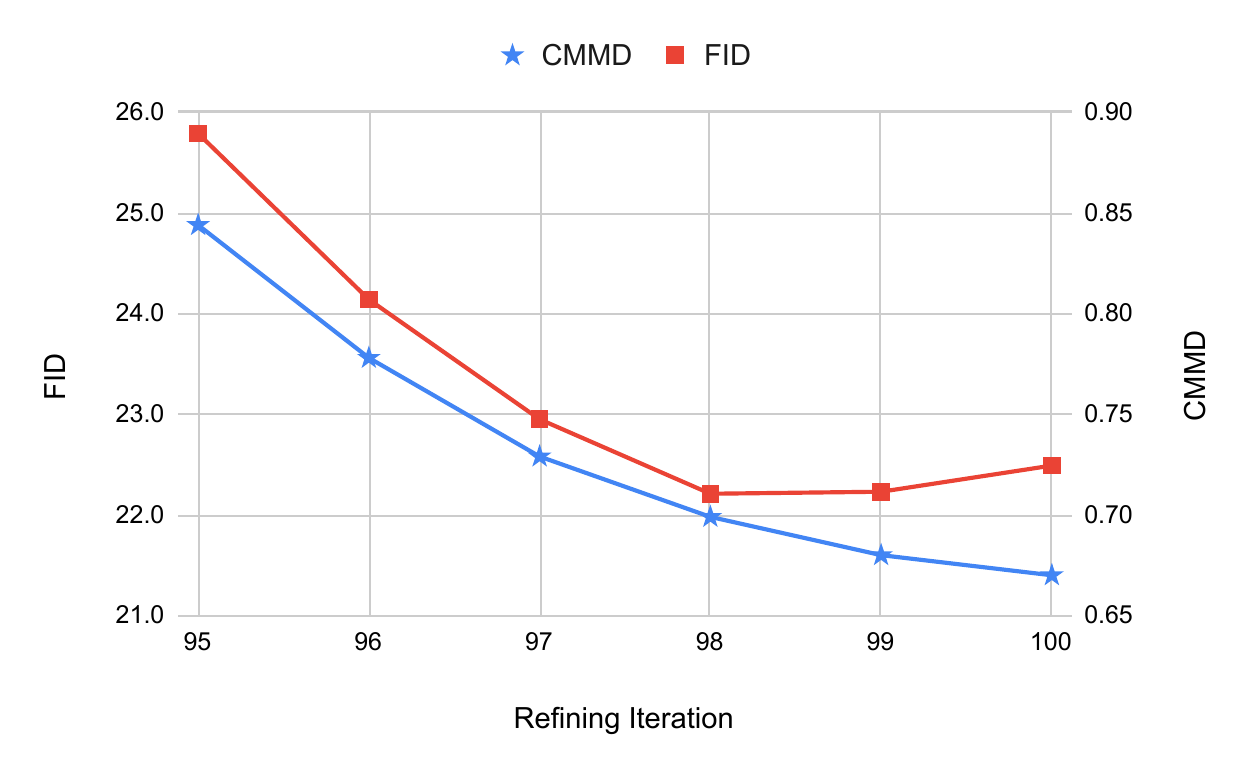}
\caption{{\it  Behavior of FID and CMMD for StableDiffusion steps. CMMD monotonically improves (goes down), reflecting the improvements in the images. FID's behavior is not consistent, it mistakenly suggests a decrease in quality in the last two iterations.\vspace{-0.5cm}}}
\label{fig:sd_iterations}
\end{figure}

Figure~\ref{fig:sd_iterations} shows an evaluation of the last 5 iterations of a 100-iteration Stable Diffusion model. Our proposed CMMD metric monotonically improves (decreases) with the progression of the iterations, whereas FID has unexpected behavior. We focus on the more subtle differences in the final iterations of Stable Diffusion, since both FID and CMMD showed monotonicity at the easily-detectable high noise levels in the initial iterations.

\subsection{Image Distortions}\label{sec:image_distortions}
\label{sec:vqgan_distortions}
Here, we provide additional evidence that FID does not accurately reflect image quality under complex image distortions. It was shown in~\cite{heusel2018gans} that FID accurately captures image distortions under low-level image processing distortions such as Gaussian noise and Gaussian blur. Since Inception embeddings are trained on ImageNet images without extreme data augmentation, it is not surprising that FID is able to identify these distortion. However, in this section, we show that FID is unable to identify more complex noise added in the latent space.

\begin{figure}[t]
\centering
\includegraphics[width=0.48\textwidth,trim={1.2cm 1.8cm 1.2cm 0.5cm},clip]{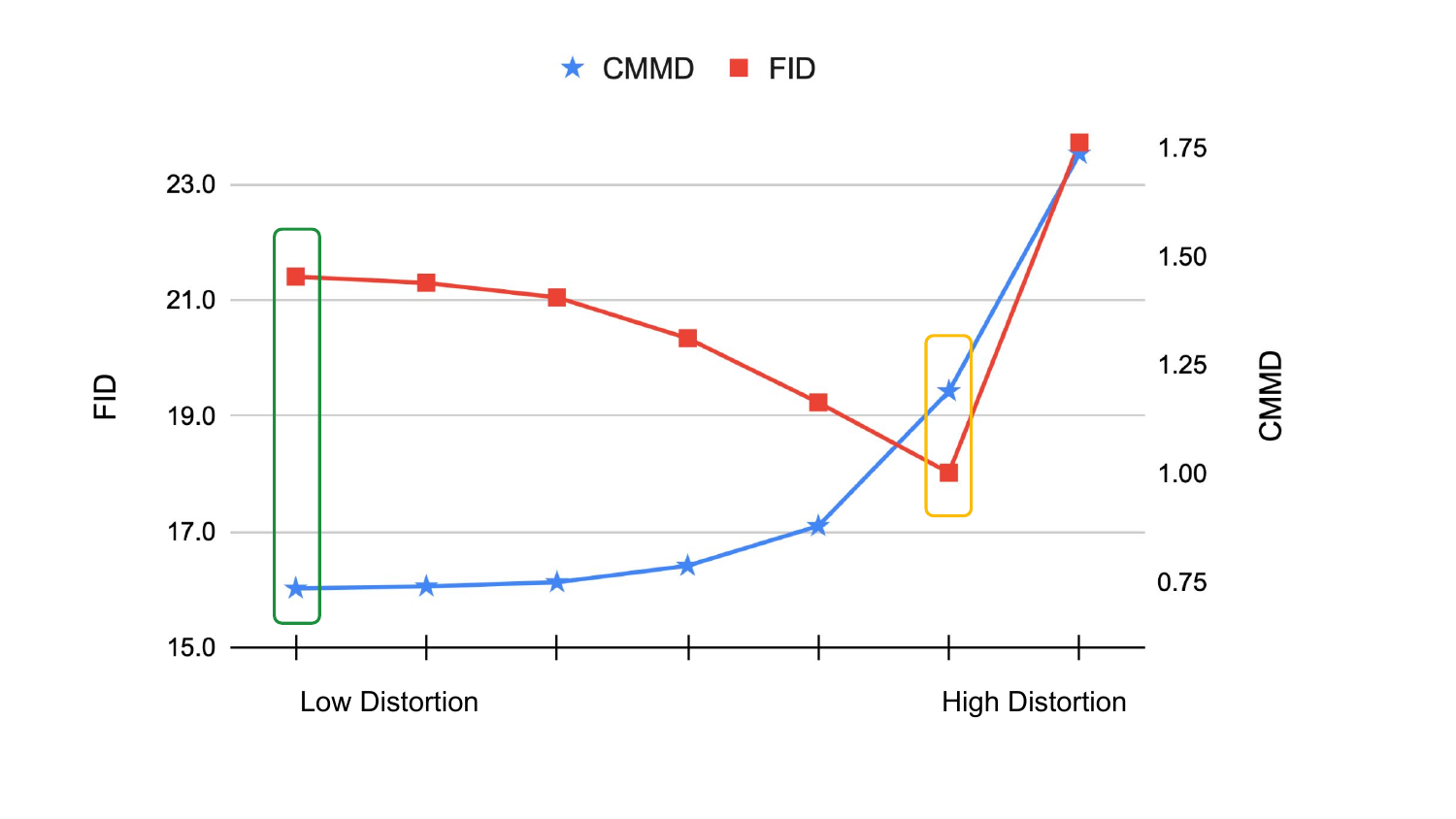}
\caption{{\it  Behavior of FID and CMMD under latent space noise added to generated images. CMMD monotonically goes up, reflecting the quality degradation of the images. FID's behavior is inconsistent, it mistakenly suggests an increase of quality. Image sets highlighted in green and yellow are visualized in Figure~\ref{fig:image_vis_muse_distortions}'s top and bottom rows, respectively.\vspace{-0.5cm}}}
\label{fig:distorted_muse_images}
\end{figure}

To this end, we take a set of images generated by Muse and progressively distort them by adding noise in the VQGAN latent space~\cite{esser2021taming}. For each image, we obtain VQGAN tokens, replace them with random tokens with probability $p$, and reconstruct the image with the VQGAN detokenizer. Example distortions are shown in Figure~\ref{fig:image_vis_muse_distortions}. The images get more and more distorted with increasing $p$ and the quality loss with increasing $p$ is visibly obvious. However, as shown in Figure~\ref{fig:distorted_muse_images}, FID fails to reflect the degradation in image quality for increasing values of $p$. Our CMMD metric, on the other hand, monotonically worsens (increases) with the distortion level $p$, correctly identifying the quality regression. Figure~\ref{fig:distorted_coco_images} shows that FID behaves poorly also when we measure the distances between progressively distorted versions (using the same procedure) of the COCO 30K dataset and the reference clean version of that dataset.

\subsection{Sample Efficiency}
\label{sec:sample_efficiency}
As stated in Section~\ref{sec:mmd}, calculating FID requires estimating a $2048 \times 2048$ covariance matrix with $4$ million entries. This requires a large number of images causing FID to have poor sample efficiency. This has also been noted by the authors of~\cite{Chong2019}. The proposed CMMD metric does not suffer from this problem thanks to its usage of MMD distance instead of the Fr\'echet distance.

\begin{figure}
\centering
\includegraphics[width=0.43\textwidth,trim={1.2cm 1.8cm 1.2cm 0.5cm},clip]{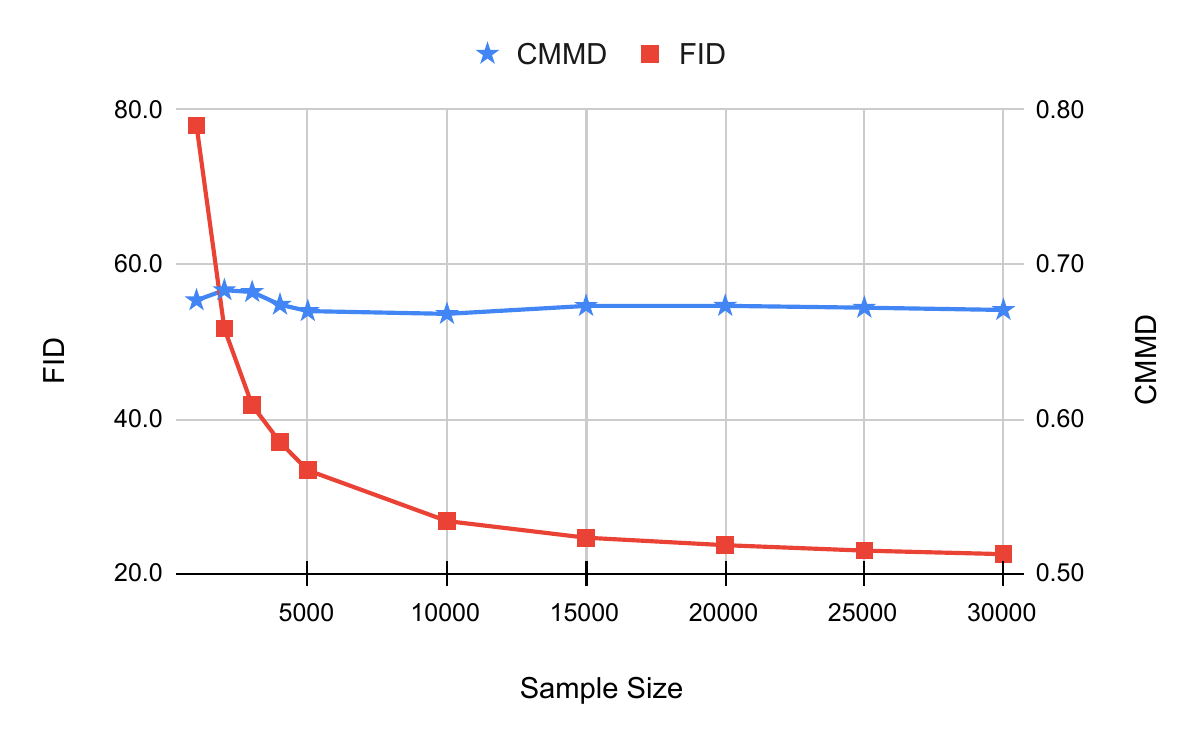}
\vspace{-0.3cm}
\includegraphics[width=0.43\textwidth,trim={1.2cm 1.8cm 0.0cm 1.5cm},clip]{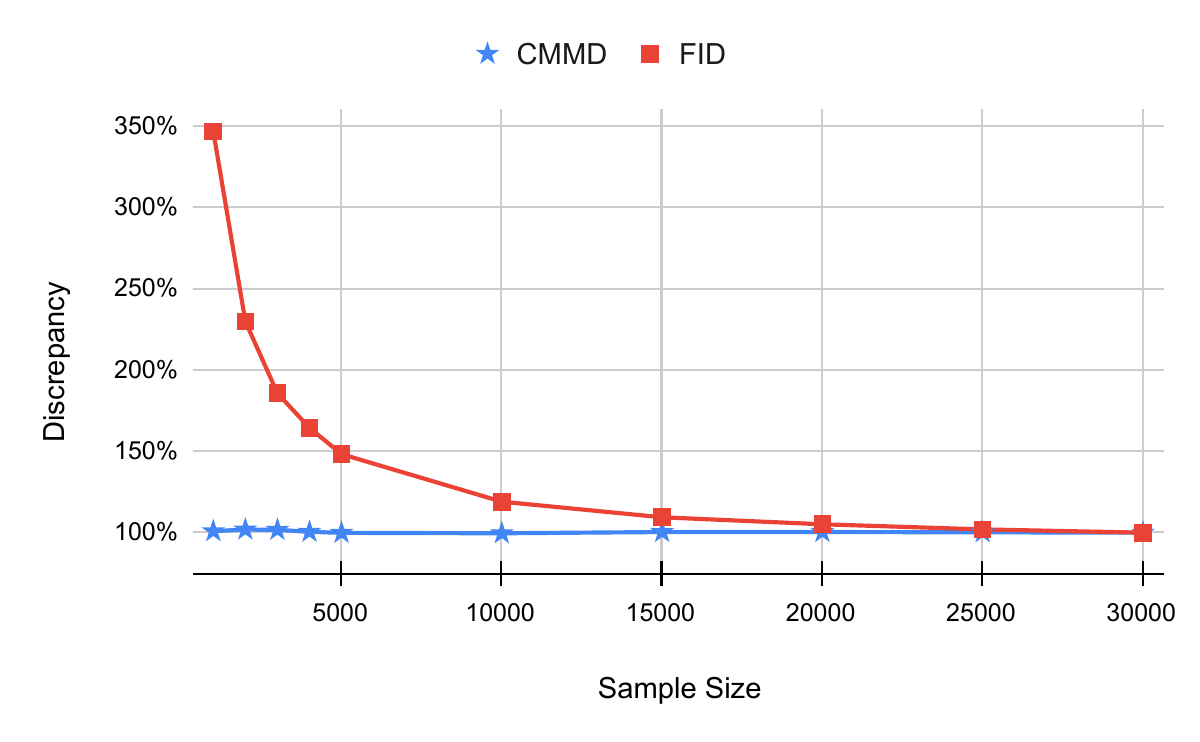}
\caption{{\it  Behavior of FID and CMMD under different sample sizes. Top: absolute values of the metrics. Bottom: Values relative to the value at $30k$ sample size.\vspace{-0.3cm}}}
\label{fig:sample_size}
\end{figure}

In Figure~\ref{fig:sample_size} we illustrate this by evaluating a Stable Diffusion model at different sample sizes (number of images) sampled randomly from the COCO 30K dataset. Note that we need more than 20,000 images to reliably estimate FID, whereas CMMD provides consistent estimates even with small image sets. This has important practical implications: development of image generation models requires fast online evaluation, e.g. as a metric tracked during training. Another relevant scenario is comparing a large number of models. Since reliable estimation of FID requires generating a large number of images, FID evaluation is costly and time consuming. In contrast, CMMD can be evaluated fast by generating only a small number of images. CMMD evaluation is faster than FID evaluation for two reasons: 1) it requires only a small number of images to be generated. 2) once the images are generated the computation of CMMD is faster than the FID computation as discussed in the next section.

\vspace{-0.2cm}
\subsection{Computational Cost}
\label{sec:computational_cost}
Let $n$ be the number of images, and let $d$ be the embedding length. The cost of computing the Fr\'echet distance (FD) is dominated by the matrix square root operation on a $d \times d$ matrix, which is expensive and not easily parallelizable. The cost of computing the unbiased version $\text{FD}_\infty$ is even higher, since it requires computing FD multiple times with different sample sizes. The asymptotic complexity of computing MMD is $O(n^2d)$. However, in practice, MMD can be computed very efficiently, since it only involves matrix multiplications which are trivially parallelizable and highly optimized in any deep learning library such as Tensorflow, PyTorch, and JAX.

Table~\ref{tab:compute_time} shows an empirical runtime comparison of computing FD and MMD on a set of size $n=30,000$ with $d=2048$ dimensional features on a TPUv4 platform with a JAX implementation. For FD calculations, we use our JAX implementation and publicly available PyTorch/numpy implementations from \cite{parmar2021cleanfid} and \cite{Chong2019} and report the best runtime. In the same table, we also report the runtime for Inception and CLIP feature extraction for a batch of 32 images.

\begin{table}[t]
    \centering
    \vspace{5pt}
    \begin{tabular}{lc}
        \toprule
        Operation           & Time \\
        \midrule
        Fr\'echet distance &   7007.59 $\pm$ 231 ms  \\
        MMD distance  &   71.42 $\pm$ 0.67 ms \\
        Inception model inference & 2.076 $\pm$ 0.15 ms \\
        CLIP model inference & 1.955 $\pm$ 0.14 ms \\
        \bottomrule
    \end{tabular}
    \vspace{5pt}
    \caption{\it Comparing runtime for computing Fr\'echet/MMD distances and Inception/CLIP feature extractions. \vspace{-0.3cm}}
    \label{tab:compute_time}
\end{table}

%% file: 05_discussion.tex
\vspace{-0.2cm}
\section{Discussion}
We encourage image generation researchers to rethink the use of FID as a primary evaluation metric for image quality. Our findings that FID correlates poorly with human raters, that it does not reflect gradual improvement of iterative text-to-image models and that it does not capture obvious distortions add to a growing body of criticism~\cite{Chong2019, parmar2021cleanfid}. We are concerned that reliance on FID could lead to flawed rankings among the image generation methods, and that good ideas could be rejected prematurely.
To address these concerns we propose CMMD as a more robust metric, suitable for evaluation of modern text-to-image models.

\section*{Acknowledgment}
We would like to thank Wittawat Jitkrittum for the valuable discussions.

%% file: appendix.tex
\appendix

\noindent
\section*{Appendix}

\section{Multivariate Normality Tests}
\label{appendix:normality_tests}
Fr\'echet Inception Distance (FID) hinges on the multivariate normality assumption. Since there is no canonical test, we show that the Inception features for a typical image dataset like COCO 30K do not satisfy this assumption using three different widely-accepted statistical tests: Mardia's skewness test~\cite{mardia1970measures}, Mardia's kurtosis test~\cite{mardia1970measures} and Henze-Zirkler test~\cite{henze1990class}.

The null hypothesis for all of the tests is that the sample is drawn from a multivariate normal distribution. Different tests use different statistics as described below.

\subsubsection*{Mardia's Skewness Test}

For a random sample of $\vx_1, \x_2, \dots, \x_n \in \mathbb{R}^d$, a measure of multivariate skewness is,
\begin{equation}
    A = \frac{1}{6n} \sum_{i=1}^n \sum_{j=1}^n \left[ (\vx_i - \bar{\vx})^T \hat{\mathbf{\Sigma}}^{-1}(\vx_j - \bar{\vx}) \right]^3.
\end{equation}
Where $\hat{\mathbf{\Sigma}}$ is the biased sample covariance matrix, and $\bar{\vx}$ is the sample mean.

Mardia~\cite{mardia1970measures} showed that under the null hypothesis that $\vx_i$s are multivariate normally distributed, the statistic $A$ will be asymptotically chi-squared distributed with $d(d+1)(d+2)/6$ degrees of freedom. Therefore, the normality of a given sample can be tested by checking how extreme the calculated $A$-statistic is under this assumption. For Inception embeddings computed on the COCO 30K dataset, this test rejects the normality assumption with a $p$-value of $0.0$, up to machine precision.

\subsubsection*{Mardia's Kurtosis Test}

For a random sample of $\vx_1, \x_2, \dots, \x_n \in \mathbb{R}^d$, a measure of multivariate kurtosis is,
\begin{align}
    B = \sqrt{\frac{n}{8d(d+2)}}\Biggl\{{1 \over n} \sum_{i=1}^n &\left[(\mathbf{x}_i - \bar{\mathbf{x}})^\mathrm{T}\;\hat{\boldsymbol\Sigma}^{-1} (\mathbf{x}_i - \bar{\mathbf{x}}) \right]^2 \nonumber \\
    &- d(d+2) \Biggl\}.
\end{align}

It was shown in~\cite{mardia1970measures} that, under the null hypothesis that $\vx_i$s are multivariate normally distributed, the statistic $B$ will be asymptotically standard normally distributed. For Inception embeddings computed on the COCO 30K dataset, this test also rejects the normality assumption with a $p$-value of $0.0$, To intuitively understand the confidence of the outcome: this Mardia's test places the test statistics $19,023$ standard deviations away from the mean in a normal distribution. This indicates the test's extreme confidence in rejecting the normality of Inception embeddings.

\subsubsection*{Henze-Zirkler Test}

The Henze-Zirkler test~\cite{henze1990class} is based on a functional that measures the distance between two distributions and has the property that, when one of the distributions is standard multivariate normal, it is zero if and only if the second distribution is also standard multivariate normal. The Henze-Zirkler test has been shown to be affine invariant and to have better power performance compared to alternative multivariate normal tests.

The Henze-Zirkler test's $p$-value for Inception embeddings of COCO 30K is again $0.0$ up to the machine precision. Therefore, the Henze-Zirkler test also rejects the normal assumption on Inception embeddings with overwhelmingly high confidence.

\section{Synthetic Experiment Details}
\label{appendix:mog_details}
In this section, we discuss the details of the experiment described in Section~\ref{sec:mog_toy_example}. As the reference distribution, we use an isotropic Gaussian distribution centered at the origin with a covariance matrix $\sigma^2\mathbf{I}_2$, where $\mathbf{I}_2$ is the $2 \times 2$ identity matrix. The second distribution consists of four different equally-likely Gaussians, centered at the coordinates $(\lambda, 0), (0, \lambda), (-\lambda, 0), (0, -\lambda)$, and each with the covariance matrix $\tau_{\lambda}^2\mathbf{I}_2$. In Table~\ref{tbl:mog_results}, we show the distribution visualizations (first row), and the behavior of different distance metrics (remaining rows) with increasing values of $\lambda$. As $\lambda$ increases, $\tau_\lambda$ is adjusted as described below so that the overall covariance matrix of the mixture-of-Gaussians distribution remains equal to $\sigma^2\mathbf{I}_2$. Trivially, the mean of the mixture-of-Gaussians is the origin. Therefore, as $\lambda$ varies, both the mean and the covariance matrix of the mixture-of-Gaussians distribution remain equal to the reference distribution. Therefore, both FD and $\text{FD}\infty$ estimated using Eq.~\ref{eqn:fd_for_gaussians} remain zero as $\lambda$ increases. This is obviously misleading as the mixture-of-Gaussians distribution gets further and further away from the reference as $\lambda$ increases. This error is a direct consequence of the incorrect normality assumption for the mixture-of-Gaussians distribution.

To see the relationship between $\tau_\lambda$ and $\lambda$ that keeps the overall covariance matrix equal to $\sigma^2\mathbf{I}_2$, consider a mixture distribution consisting of 1-D PDFs $f_1, f_2, \dots, f_n$ with weights $p_1, p_2, \dots, p_n$, where each $p_i > 0$ and $\sum_i p_i = 1$. The PDF of the mixture distribution is then given by $f(x) = \sum_i p_i f_i(x)$. It follows from the definition of the expected value that, $\mu^{(k)} = \sum_i p_i \mu_i^{(k)}$, where $\mu^{(k)}$ and $\mu_i^{(k)}$ are the $k^\text{th}$ raw moment of $f$ and $f_i$, respectively. Recall also that variance is $\mu^{(2)} - \{\mu^{(1)}\}^2$. By applying the above result to $x$ and $y$ coordinates individually, we see that the overall covariance matrix of the above mixture of four Gaussians, when they are away from the mean by $\lambda$, is given by $(\tau_\lambda^2 + \lambda^2 / 2)\mathbf{I}_2$. Setting $\tau_\lambda^2 = \sigma^2 - \lambda^2/2$ therefore keeps the overall covariance matrix at $\sigma^2\mathbf{I}_2$ as we vary $\lambda$.